Paper ID 290

# Virtual passengers for real car solutions: synthetic datasets

**Paola Natalia Cañas[*], Juan Diego Ortega, Marcos Nieto, Oihana Otaegui**
{pncanas, jdortega, mnieto, ootaegui}@vicomtech.org, Vicomtech, Spain

**Abstract**

Strategies that include the generation of synthetic data are beginning to be viable as obtaining real data can be logistically complicated, very expensive or slow. Not only the capture of the data can lead to complications, but also its annotation. To achieve high-fidelity data for training intelligent systems, we have built a 3D scenario and set-up to resemble reality as closely as possible. With our approach, it is possible to configure and vary parameters to add randomness to the scene and, in this way, allow variation in data, which is so important in the construction of a dataset. Besides, the annotation task is already included in the data generation exercise, rather than being a post-capture task, which can save a lot of resources. We present the process and concept of synthetic data generation in an automotive context, specifically for driver and passenger monitoring purposes, as an alternative to real data capturing.

**Introduction**

The progress of the automotive industry and the interest in the development of autonomous vehicles has made this field one of the main areas of application of deep learning. In order to achieve a SAE-L3/L4 level of autonomy [1], an effort is not only required to develop Advanced Driver Assistance Systems (ADAS), but also algorithms that allow the car to know the trajectory it should take, traffic signs recognition, other road entities detection; in other words, exterior perception algorithms. It is also necessary for the vehicle to be aware of the state of the passengers or to have in-cabin perception algorithms that constitute a passenger/driver monitoring system. The functionalities to be implemented can be driver distraction detection, drowsiness detection, gaze estimation, occupancy estimation, among others. Some functionalities will focus on the driver and others on the passengers in general. In this paper, we present a use case for general passenger monitoring.

Depending on the objective or the complexity of the task to be carried out, the dataset may require great effort for its creation in a real context. Also, the annotation can take a lot of time and work to complete. That is why, given the lack of data and considering the possible obstacles that the recording and annotation of a real dataset imply, we present an alternative: the creation of synthetic data for interior perception algorithms, expecting that it will offer the following advantages:
- Implementation of different and personalized types of sensors.
- Automatic ground truth creation.
- Multiple labels generation.
- The capture of the same scene from different positions or angles.

Virtual passengers for real car solutions: synthetic datasets

- Easy lightning conditions variation.
- No consideration of sensitive personal data policies or privacy agreements.
- No volunteer coordination.
- No vehicle or real cameras are required.
- No annotation errors.

It is a thing of the past that the generation of synthetic data is difficult or unfeasible compared to the capture of real data. More and more tools have been developed or improved that facilitate this process. In an attempt to parameterize characteristics and create algorithms that allow their variation and the mass production of 3D models of acceptable quality, although without the realism that could be provided by detailed modelling such as the one done for artistic purposes (films, video games, etc), this data generation technique has started to gain devotees and invited many researchers to consider this option for computer vision solutions.

**State of the art**

Most of the current work is related to the vehicle's exterior, while little is found for the interior. This can be seen in the number of datasets available for the development of driver or passenger monitoring systems, in contrast to what can be found for those dedicated to the vehicle exterior.

Large and well-known datasets for exterior perception are pillars for research in this field, such as KITTI [2], a dataset launched in 2012 that contains different types of data like optical flow, visual odometry and 2D/3D object annotations for road actors detection; nuScences [3], a multimodal dataset for autonomous driving with 15 hours of annotated HD video, 1.4 billion lidar points and nearly 100,000 images; among other datasets. While for inside-cabin perception, dataset efforts have been made individually, and as a result, there are fewer in number, small in size and task-specific datasets. For example, datasets for driver distraction detection [4,5], drowsiness detection [6,7], head-pose estimation [8], etc. More recently, the DMD (Driver Monitoring Dataset) [9] was released and includes data for four specific detection tasks: distraction, drowsiness, gaze and hands-wheel interaction. However, despite being a robust dataset covering more than one task, it is centred on the driver.

Regarding synthetic datasets, few have been created for the automotive industry. For exterior applications, KITTI has a synthetic version called virtual KITTI [10]; it was generated with the 3D modelling software Unity[1] and offers ground truth for object detection, tracking, scene and instance segmentation, depth, and optical flow. There is also nuReality[11], an open-source virtual environment designed to test the interactions between pedestrians and autonomous vehicles.

The main reference of these types of datasets dedicated to vehicle-interior perception systems is SVIRO [12]. It was developed with Blender[2] and is planned to support rear seat occupancy estimators, offering RGB, simulated infrared and depth images; it contains bounding boxes for object detection, masks for semantic and instance segmentation and key points for pose estimation for each synthetic scenery. Another similar and more recent dataset is TICam [13].

Outside the automotive domain, there are other examples of synthetic datasets, some of which can also be used in this context, such as SynthEyes [14], which has 11M synthetic close-up eye images for a wide range of head poses, gaze directions, and illumination conditions. Another important consideration is the human data generator PeopleSansPeople from Unity [15]. It contains 3D human assets, a parameterized lighting and camera system, and generates 2D and 3D bounding boxes,

---
[1] https://unity.com/
[2] https://www.blender.org/

instance and semantic segmentation, and COCO pose labels.

**Synthetic data generation**

A high-fidelity dataset consists of data generated as a result of a stochastic process that is realistic or close enough to reality and, in computer vision applications, useful for algorithm training. The creation process presented in this document adds randomness to its variables and has a level of realism close to reality that, if needed, can be compensable with domain adaptation or fine-tuning techniques. In addition, the automatic generation of labels in this process results in accurate annotations. It does not allow confusion because of the subjectivity of the annotators or errors of an automatic annotation process, contributing to the fidelity of the dataset.

Complying with the ethical bases of the recent European regulation on artificial intelligence, the generation of synthetic data is in line with many of the proposed statutes. Under the concept of Trustworthy AI, a series of guidelines have been defined to ensure the proper and fair development, application and use of artificial intelligence by humanity [16]. Specifically, synthetic data goes hand in hand with the following ethical rules: privacy, where no biometric or sensitive data of any subject is included in the data (since they are all synthetic); avoidance of unfair bias, this can be accomplished since the variables on which the generation of the models depends are parametrised, by adjusting the appropriate parameters and adding randomness, existing bias can be eliminated; data provenance is another ethical rule that is fulfilled, since this is a "controlled" creation process, in the sense that the steps and variables that are part of the creation of a synthetic data are known and defined, the origin of the data can be determined, offering transparency and also reproducibility properties. Despite being a stochastic process, random variables can have a common seed that allows reproducibility, explainability and, at the same time, sufficient differentiation between data to achieve the desired variation and bias avoidance.

High-fidelity data that contributes to the development of artificial intelligence algorithms that follow ethical guidelines is the data that industry and academia will adopt. The synthetic creation of this type of data is what we intend to describe. Particularly, it might be suitable for the creation of datasets dedicated to interior perception algorithms or for any field. The following are key points to have in mind when building a synthetic dataset:

*Engine for 3D scenario*

Nowadays, the advances in computational power and dedicated 3D engines enable us to render 3D scenarios with excellent fidelity and with everyday computers. Based on state of the art, the most common ones and all-purpose software are Blender and Unity.

Blender is an open-source software to create 3D scenarios, and it includes Python to automate and control many tasks through commands. This means that it is possible to make a script to create and configure objects or scene elements in the 3D world.

Unity is a video game engine where 3D modelling can be done as well, and it has a free and paid version (Unity Pro). It is compatible with Blender and many assets from other 3D software. There is also the possibility to give instructions through a script; in this case, it has to be done with C# o Boo (that has a similar syntaxis as Python).

*Human & object models*

The objects imported into the scene will depend on the use case of the dataset, as well as their level of realism and their variation in appearance. These objects can be modelled from scratch in the same software; depending on the 3D modeller's expertise, time and resources, the object may or may not meet expectations with its level of detail. Creating objects from scratch is always an option; however, there are 3D model markets where these assets can be obtained, some free, others with restricted or paid use. Some of these markets are:
- Blender Market (https://blendermarket.com)
- Hum3D (https://hum3d.com)
- CGTrader (https://www.cgtrader.com)
- Sketchfab store (https://sketchfab.com)
- Free 3D (https://free3d.com)

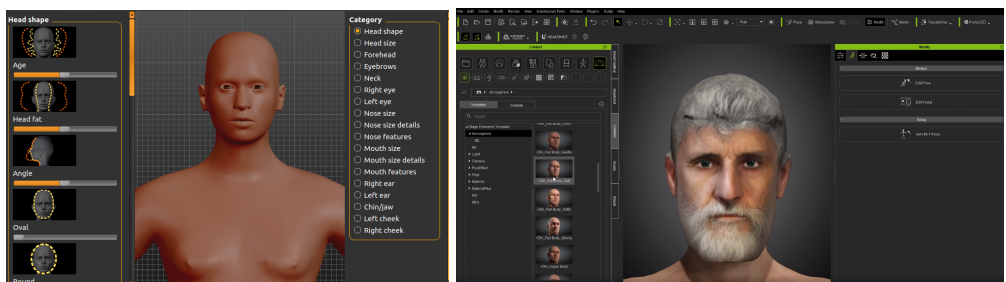

**Figure 1 - Makehuman and Character Creator user interface.**

As for humans, it is more complex. The availability of human models is usually scarce due to the complexity involved in making a human with a high level of detail (body and facial design, clothing, etc.). Purchasing a pack of human models can be considered, but these are usually expensive and have a limited number of humans. Depending on the use case, these human packs may be enough. However, there are software solutions dedicated to building characters because this is a problem faced not only by a computer vision researcher but also by an animator of animated series, a video game designer or any other person who requires scenes with people. This software come to facilitate this arduous creation of people, doing all the work of creating curves and shapes that make up the person's physiognomy and leaving the user the option to vary subtly and in a configurable way many parameters that come to define and distinguish one person from another. Some of those character creation programs are:

- *Makehuman[3]:* This software is open source and offers the possibility to create virtual humans using a graphical user interface. Humans are created through the manipulation of controls that allow for the blending of different human attributes to create unique 3D human characters. This is mostly used with Blender but models can be exported in other formats compatible with many 3D engines. The level of realism is not high but could be enough for some use cases. It has a functionality of mass production that allows random generation of humans without going through the process of designing one by one. The community can design assets like clothing and include them in the creation of humans.

- *Character Creator[4]:* This product of Reallusion company is a paid software that allows users to easily create and customize highly-realistic characters and import them into a scene from Unity or Blender (or many more). It also has a stock of assets like clothing that can be implemented in human models. However, it doesn't have the option of random creation of characters as Makehuman. Along with Character Creator, Reallusion offers other compatible products that can complement human production if needed. Such as iClone to animate or create a virtual version of a real human. This tool is mostly used with Unity, but it can be compatible with Blender as well.

---

[3] http://www.makehumancommunity.org/
[4] https://www.reallusion.com/character-creator/

Virtual passengers for real car solutions: synthetic datasets*Lighting & Background*

The lighting of the scene is given by a set-up of lights or by the lighting provided by the background of the 3D world that is established. The first lighting type can include the following lighting types that are common to both 3D engines (Blender and Unity). Randomness in lighting can be given by varying the type of lights, intensity, position, direction, etc. Also, many sets or lighting configurations can be established, and, for each sample, a random one can be chosen.

- *Point light:* It is positioned at a point in the scene and radiates equal intensity in all directions.
- *Spot light:* This one has a cone-shaped light emission and is located at a point in the scene.
- *Directional light (Unity) or sun light (Blender):* This type of lighting emits constant intensity in a single direction from an infinite far away distance.
- *Area light:* It simulates a rectangle-shaped surface light, emitting intensity from one side uniformly in all directions.

The second type of lighting is included when an HDRI (High Dynamic Range Image) is set as the background of the 3D world. These photos are 360º images that adapt to the background of the scene and include lighting related to the environment it represents. The largest repository of these images in high quality is Polyhaven[5]. There can be found different backgrounds from cities, room environments, nature, among others. Some examples are presented in Figure 2.

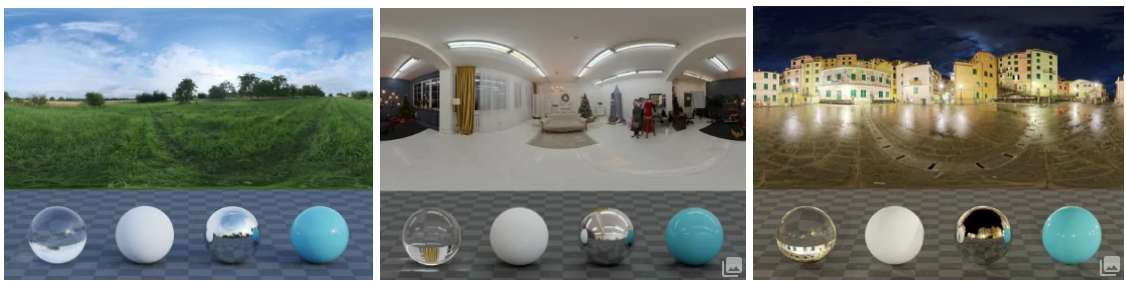

**Figure 2 - HDRI's examples: nature, room and city environments**

*Camera type & positioning*

The position and type of sensors are one of the advantages of synthetic data. This is because putting an extra camera does not imply any additional cost other than generating renders from a new perspective. Unity and Blender have implemented different types of cameras, from fisheye to orthographic, offering the configuration of features such as resolution, the field of view, focal length, among others. This allows the location of one or more cameras in the scene, obtaining images with the same distortion that the real camera would have.

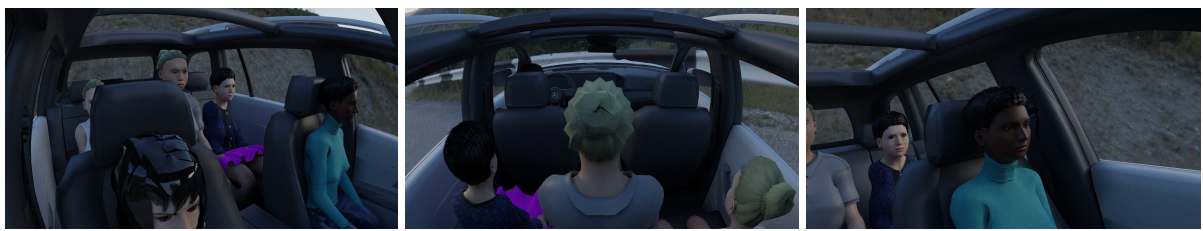

---

[5] https://polyhaven.com/hdris

**Figure 3 - Examples of camera positioning and angle**

In addition, it has been demonstrated that it is possible to simulate sensors that capture depth information and IR cameras as described in [12,13].

*Labelling*

Within the possibilities of annotations, there are the most common labels for computer vision tasks. These are semantic and instance annotation, bounding boxes, distance to the camera (depth information); as well as other ones that can be derived from the information of the 3D software, such as the location and/or rotation of a person or object in the scene, distance from them, etc. This also counts for bone rotation (in a human skeleton), which could lead to labels of head pose, gaze direction, body pose, etc. or bone position, with which body landmarks could be established.

Data being annotated directly with information from the 3D software ensures that the labels are precise and eliminates the arduous work of an annotation task.

**Description of interior perception use case**

Our approach towards the creation of synthetic data has been through the creation of a dataset to train and validate in-cabin occupancy estimation algorithms. Here, we will describe the methodology and technologies used to create this dataset.

*Technical choices*

Our scene is composed of a car and virtual passengers inside it. The car model, shown in Figure 4, was obtained in CGTrader and is a MercedesML free 3D model. We chose this model as it offers a good level of detail in the visual features of the interior of a car.

We chose to use Blender along with Makehuman to prepare the 3D world due to its compatibility, ease of use, Python scripting, and mass production option from Makehuman. This mass production plugin creates new characters from the modification of a base human (which can be set). These modifications include aspects such as height, width, proportions of the body parts, the size of the eyes, mouth, forehead, etc. It has some assets of clothing, hair types and accessories by default, but contributions from the community can easily be found. Several distinct characters can be automatically and randomly generated; their variations obey the configuration of ranges of physical aspects or variables, such as those mentioned above. For example, setting the minimum and maximum of height will assign a random height within that range to the generated characters. Assets like clothing and hair are also random.

In addition, a skeleton must be added to them to make humans adopt a posture (sitting in this case). Makehuman allows generated characters to have a skeleton supported by the CMU Graphics Lab Motion Capture Database [17]. This enables later character pose variation by modifying the position and rotation of the skeleton bones.

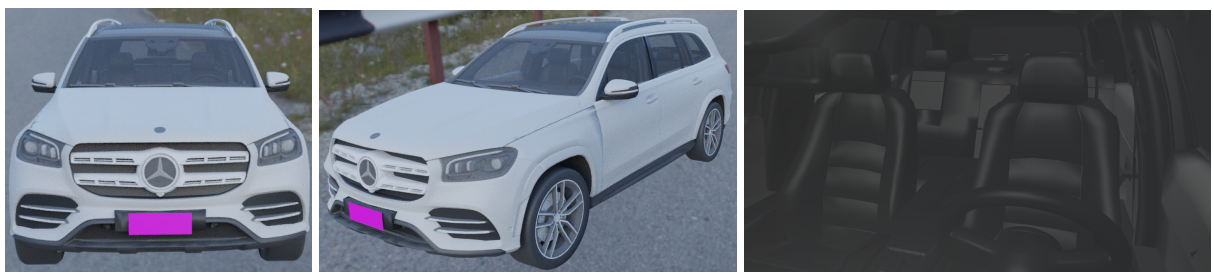

**Figure 4 - Car 3D model with interior modelling**

For the lighting, HDRI's were used as scene background, and the illumination it included was the source of light of the scene. A stack of these HDRI pictures was collected to import to the scene randomly.

In order to get all the passengers properly in the image frame, the camera position must be set on the centre mirror pointing towards the inside of the cabin. After evaluating various camera setups in Blender, the following characteristics were established, as can be appreciated in Figure 6, a panoramic camera of a Fisheye Equisolid type, with a field of view of 180º and a sensor width of 5.3mm.

*Methodology*

As an individual result, we want to obtain an RGB image with masks or semantic segmentation of the passengers and their respective 2D bounding boxes. For this, the process represented in Figure 5 is followed. First, humans are created in Makehuman. As mentioned above, some ranges of physical characteristics are configured to generate the models randomly. In this way, 30 humans were generated. Then, the car model is added to the scene, and the Blender environment is prepared so that it can execute a script. This script takes into account variation ranges that are previously defined, among which are: the position of the passengers (to occupy a car seat), the range of movement of each bone that was supposed to move; for example, the neck could vary its position vertically (up and down) and horizontally (right and left) in about 15º in both directions. These ranges must be carefully defined to achieve natural poses.

This variation of the characters, the rendering of the RGB images, and the masks must be automated, so the system produces a new individual sample of the dataset each time. For this, a script that follows the instructions inside the grey rectangle that says "iteration" was created. The script randomly chooses five human models and imports them into the scene, placing them into a seat. After that, it changes their posture by rotating some of their bones within ranges of variation. Taking one random HDRI from the collected stack changes the background and the lighting of the scene.

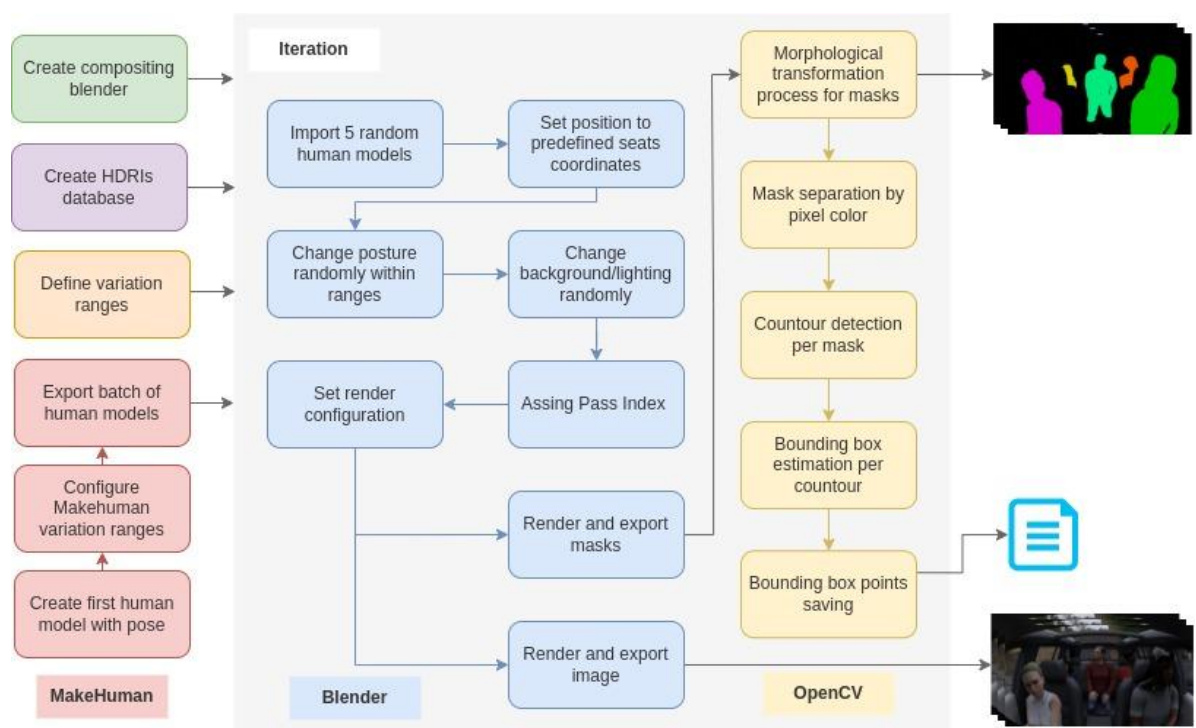

**Figure 5 - Synthetic data generation process**

Virtual passengers for real car solutions: synthetic datasets

To make semantic segmentation, each human model must be assigned with a property in Blender called Pass Index. Later in the render configuration, this index will help differentiate the pixels that belong to each person, adding colour to those pixels to create the masks. However, what seems to be a texture problem causes some noise (tiny holes) in the final mask image, which is why a morphological transformation with OpenCV must be performed. A Closing (Dilation followed by Erosion) operation was applied to the mask images to eliminate the noise.

The bounding boxes are obtained from the masks through contour identification operations. The mask images were filtered by colours to get each individual person's mask. For each person's mask, a find contours function from OpenCV was applied. This outputs a set of points approximated to a polygon, which is then approximated to a rectangle. This final rectangle is our bounding box (Figure 6). The coordinates (x,y), the width and height of each person's bounding box in the picture is saved in a text file. As a result, we have an RGB image, a mask image and a text file with data about bounding boxes.

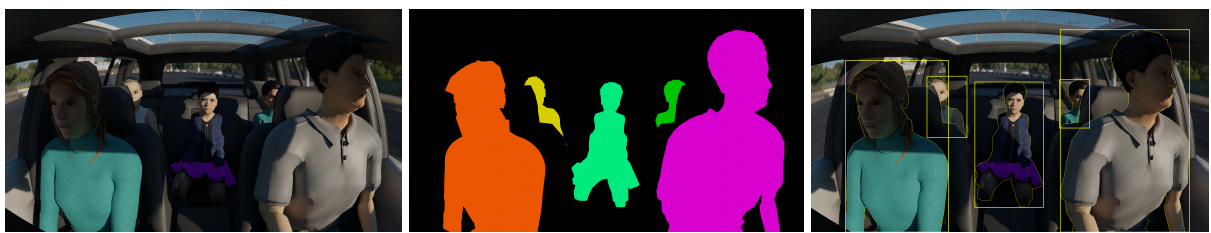

**Figure 6 - Outputs of the data generation. RGB image, masks and bounding boxes.**

**Conclusions**

We have presented a description of the requirements and options to make a synthetic dataset for interior perception algorithms. We also propose a methodology defined from our own experience in building a synthetic dataset, which consists of the preparation of a 3D environment and a script that, through iterations of itself, new samples are created. Therefore, we had automated a pipeline for the generation of annotated samples, achieving a different result every time it is executed. That allowed us to create a validation dataset and gives us the opportunity to generate more data in the future if needed.

Synthetic data is a reality within the computer vision world. It can be seen that big and standard used datasets also have their synthetic version. We believe this will still be a good alternative for building datasets since tools are moving towards developing a rapid and user-friendly creation and customization of 3D models and scenarios.

The creation of synthetic data can speed up the development process, which can run in parallel to sensor configuration and set-up, creating the annotations along with the visual material, reducing time-to-market and thus saving time and costs. For this and all the advantages presented in this document, the inclusion of this type of data in deep learning algorithms must be considered nowadays.